\documentclass[conference]{IEEEtran}

\usepackage{cite}
\usepackage{array}
\usepackage{amsmath,amssymb,amsfonts}
\usepackage{algorithmic}
\usepackage{graphicx}
\usepackage{textcomp}
\usepackage{stfloats}
\usepackage{xcolor}
\usepackage{hyperref}
\usepackage{booktabs} 
\usepackage{balance}
\usepackage{subcaption}

\def\BibTeX{{\rm B\kern-.05em{\sc i\kern-.025em b}\kern-.08em
    T\kern-.1667em\lower.7ex\hbox{E}\kern-.125emX}}
    
\begin{document}

\title{Goal‑Conditioned Reinforcement Learning for Data‑Driven Maritime Navigation} 

\author{%
\centering
\IEEEauthorblockN{Vaishnav Vaidheeswaran, Dilith Jayakody, Samruddhi Mulay}
\IEEEauthorblockN{Anand Lo, Md Mahbub Alam, Gabriel Spadon}
\IEEEauthorblockA{\textit{Faculty of Computer Science}, 
\textit{Dalhousie University}, \textit{Halifax}, \textit{Canada} \\
\{vaishnav, dilith, samruddhi, anandlo, mahbub.alam, spadon\}@dal.ca}
}

\maketitle

\begin{abstract}
Routing vessels through narrow and dynamic waterways is challenging due to changing environmental conditions and operational constraints. Existing vessel-routing studies typically fail to generalize across multiple origin–destination pairs and do not exploit large-scale, data-driven traffic graphs. In this paper, we propose a reinforcement learning solution for big maritime data that can learn to find a route across multiple origin--destination pairs while adapting to different hexagonal grid resolutions. Agents learn to select direction and speed under continuous observations in a multi-discrete action space. A reward function balances fuel efficiency, travel time, wind resistance, and route diversity, using an Automatic Identification System (AIS)‑derived traffic graph with ERA5 wind fields. The approach is demonstrated in the Gulf of St. Lawrence, one of the largest estuaries in the world. We evaluate configurations that combine Proximal Policy Optimization with recurrent networks, invalid-action masking, and exploration strategies. Our experiments demonstrate that action masking yields a clear improvement in policy performance and that supplementing penalty‑only feedback with positive shaping rewards produces additional gains.

% Dynamic vessel characteristics, including fuel consumption and variable weather, are incorporated to capture realistic operating conditions. 
\end{abstract}


\section{Introduction}

As the maritime sector grows, it creates pressing safety and environmental challenges, particularly regarding carbon emissions and the impact of transportation on the blue economy. Efficient and safe vessel routing through narrow, dynamic waterways remains a longstanding challenge due to environmental variability (e.g., wind, waves, and currents), complex coastal geometry, and operational constraints (e.g., speed limits, fuel budgets, and ETA windows) that must be respected in real time~\cite{sharif2023maritime}. Conventional marine guidance systems typically decouple guidance, control, and path planning, relying on Line-of-Sight or Proportional-Integral-Derivative (PID)- based autopilots atop precomputed waypoints, often under simplified dynamics or calm-water assumptions. However, such pipelines can struggle to adapt to time-varying disturbances, to trade off multiple competing objectives (e.g., fuel \textit{versus} time \textit{versus} risk), and to generalize across regions without manual retuning~\cite{deraj2023deep}.

The decision-making models for ship navigation span a wide range of techniques, including expert systems, fuzzy logic, genetic algorithms, dynamic programming, neural networks, and, more recently, deep reinforcement learning (DRL) ~\cite{alamoush2025maritime}. Rule- and knowledge-based systems offer interpretable frameworks aligned with the International Regulations for Preventing Collisions at Sea (COLREGs), but are less effective in dynamic maritime settings where environmental disturbances and variable ship behaviors prevail. Optimization methods and neural networks have enabled data-driven route planning directly from sensor and trajectory inputs; however, they often demand a high computational cost and lack transparency~\cite{zhang2019decision}.

Reinforcement Learning (RL) has emerged as a powerful alternative for addressing these challenges. Unlike traditional techniques that require explicit modeling of system dynamics (e.g., search algorithms, control, and rule-based systems), RL algorithms learn adaptive policies by interacting with the environment, making them particularly suitable for uncertain scenarios~\cite{zhang2025adaptive}. However, current RL-based approaches in maritime navigation path planning remain limited by simplified environmental models, a restricted geographic scope, and a narrow formulation that typically optimizes for a single origin–destination pair, without configurable frameworks for systematic evaluation across diverse scenarios. Goal-Conditioned Reinforcement Learning (GCRL) is emerging as a promising approach for learning policies that can adapt and generalize across multiple goals~\cite{DBLP:journals/corr/abs-1802-09464,liu2022goal}. To unlock its potential in the maritime domain, it is essential to leverage big data, particularly AIS records, which capture large-scale vessel trajectories under real-world operational conditions.

AIS datasets provide real-time tracking data on vessel positions, speeds, and courses, making them essential for large-scale monitoring and transport analytics~\cite{stach2023maritime}. These datasets enable detailed spatiotemporal analysis of vessel movements. However, the volume and variability of AIS data still demand advanced data-driven methods to extract structured knowledge and support modeling~\cite{michaelides2020decision,dalaklis2023opportunities}. In particular, using such data for training RL agents remains challenging, as historical AIS records are not inherently designed to capture interactive decision-making processes. Bridging this gap requires transforming large volumes of data into representations suitable for learning. Deep learning can extract complex spatiotemporal patterns from AIS trajectories, while RL provides a framework for learning adaptive decision-making policies. Accordingly, this paper contributes the following:

\begin{enumerate}
    \item Proposes a configurable RL formulation for maritime routing on hexagonal lattices. 
    \item Integrates ERA5 reanalysis winds to emulate realistic, time-varying conditions.
    \item Implements safe RL mechanisms, including action masking and termination conditions (in a non-masking setting).
    \item Incorporates AIS-derived vessel traffic graphs, allowing agents to learn from structured patterns and align reward choices with frequently visited cells.
    \item Designs a GCRL environment that enables agents to learn routing policies across multiple origin--destination pairs within an evaluated region and resolution.
\end{enumerate}

The remainder of this paper is organized as follows. Section~\ref{sec:literature} reviews related work on maritime navigation, AIS data analytics, and reinforcement learning. Section~\ref{sec:methodology} explains the methodology, including data processing, spatial discretization, and the design of the reinforcement learning environment. Section~\ref{sec:evaluation} presents the experimental setup and evaluation results, analyzing how different algorithms and safety mechanisms affect routing performance. Section~\ref{sec:conclusion} concludes with key remarks and outlines directions for future research.

\section{Literature Review}
\label{sec:literature}

The integration of big data and RL in maritime navigation has enabled significant advances in research, driven by the increasing availability of large-scale vessel tracking datasets and the need for autonomous route optimization~\cite{gao2022mass}. This section reviews the evolution from traditional approaches to advanced data-driven RL methods, with particular emphasis on spatial representation schemes, environmental data integration, and the scalability challenges that current maritime RL frameworks attempt to address.

\subsection{Traditional and Deep Learning for Maritime Routing}

Maritime route optimization has traditionally relied on deterministic algorithms such as Dijkstra's and A* search, along with heuristic techniques focused on single objectives like distance minimization~\cite{bi2024artificial}. While methods such as Kalman filtering provide probabilistic state estimation~\cite{drapier2024enhancing}, they struggle with the non-linear dynamics of real environments, and classical regression or machine learning models fall short in capturing spatial-temporal dependencies. Recent advances in Deep Learning have transformed trajectory prediction, with LSTMs capturing temporal movement patterns~\cite{zhang2023research}. Transformers, such as TrAISformer, have achieved high-accuracy sequence modeling from large AIS datasets~\cite{nguyen2024transformer}. More recent work has employed probabilistic feature fusion and autoencoders to improve long-term multi-path vessel trajectory forecasting~\cite{spadon2024multi}.

\begin{table*}[b]
\centering
\caption{Comparison of Maritime RL Approaches and Data Integration Methods}
\label{tab:comparison}
\resizebox{\textwidth}{!}{%
\begin{tabular}{p{1.8cm} l l l l l}
\toprule
\textbf{Attribute} & \textbf{Ours} & \textbf{ATRN}~\cite{zhang2025adaptive} & \textbf{Weather Routing}~\cite{latinopoulos2025marine} & \textbf{East Asian Routes}~\cite{kim2024advancing} & \textbf{DQN Maritime}~\cite{alam2023ai} \\
\midrule
Approach & PPO/Action-Masking + LSTM & PPO+LSTM & DDQN/DDPG & DQN/PPO & DQN \\
Spatial repr. & H3 Hexagonal & Continuous & Orthogonal grid & Grid-based & Waypoint \\
AIS data & Historical traffic & No & No & No & No \\
Weather & ECMWF ERA5 reanalysis (hourly) & Synthetic & ECMWF ERA5 reanalysis (hourly) & None & No \\
Scale & Large-scale & Limited & Medium & Regional & Limited \\
Fuel model & Cubic law & Basic & XGBoost & No & No \\
Action space & Multi-discrete & Continuous & Discrete/Cont. & Discrete & Discrete \\
GCRL & Yes & No & No & No & No \\
Safety & Action masking & COLREGs & Basic & Traffic Separation Scheme & Basic \\
Configurable & Yes & No & Partial & No & No \\
\bottomrule
\end{tabular}%
}
\end{table*}

\subsection{Reinforcement Learning in Maritime Navigation}

The application of DRL to maritime navigation is being actively explored due to its ability to learn adaptive policies through trial-and-error learning. Deep Q-Network (DQN) applications have demonstrated effectiveness in maritime collision avoidance and basic path following~\cite{alam2023ai}, though these early approaches typically operate in simplified environments with limited state representations. Recent work by Kim et al.~\cite{kim2024advancing} applied DQN and PPO to maritime route optimization in East Asian waters, demonstrating that PPO outperforms DQN in complex coastal environments through stable policy updates and better handling of continuous state-action spaces. To model the operational area for their agents, they employed a conventional square grid-based discretization of the state space. However, the use of hexagonal grids in RL for maritime navigation remains underexplored, even though hexagonal space discretization offers uniform neighbor connectivity, reduces directional bias compared to square grids, and provides a more consistent representation of movement across the ocean surface.

Recent developments have introduced more sophisticated RL architectures that better address the complex temporal and environmental dynamics in maritime navigation. Zhang et al.~\cite{zhang2025adaptive} proposed Adaptive Temporal Reinforcement Learning (ATRN), an approach that integrates LSTM networks with PPO to capture temporal features of maritime state spaces. The ATRN model incorporates comprehensive environmental factors, including wind, waves, and currents, using the Nomoto equation. The system achieved a 30\% improvement in reward values compared to PPO, DDPG, and A3C algorithms, highlighting the importance of temporal modeling in maritime RL. However, the study does not incorporate real-time weather data, limiting its adaptability to dynamic weather conditions. Furthermore, training used fixed start–end pairs, which focused each agent on a specific route rather than supporting multi-goal behavior within a single policy.

Latinopoulos et al.~\cite{latinopoulos2025marine} developed a comprehensive weather routing system using DDQN and DDPG algorithms. Their approach introduces a two-phase optimization framework based on offline pre-training using initial weather forecasts and online adaptation with real-time weather updates. DDPG achieved 12.6\% fuel consumption reduction by integrating real-time ECMWF weather data and employing XGBoost for fuel consumption prediction. However, this two-phase, route-specialized pipeline avoids learning a single policy over multiple origin–destination pairs, and does not incorporate hexagonal discretization with safety-aware neighbor masking.

Recent studies on collision avoidance in maritime traffic employ action masking techniques, such as those proposed by Krasowski et al.~\cite{krasowski2024provable}. In their approach, COLREGs are enforced during both learning and deployment by masking invalid actions, thereby ensuring that the agent selects only provably rule-compliant actions. However, to the best of our knowledge, no prior work has explored action masking methods in the context of maritime routing with RL.

Outside maritime domains, recent work combines GCRL with graph-based planning by constructing waypoint graphs from replay-buffer states. In contrast, our study relies on a fixed, data-driven traffic prior rather than learning the graph solely from agent experience~\cite{feng2025safe}.

Complementary to these safety-oriented strategies, Spadon et al.~\cite{spadon2025learning} propose constructing Markovian traffic graphs from AIS trajectory data, yielding a probabilistic description of large-scale vessel movements. While this approach is effective for capturing vessel signatures and traffic regularities, it remains primarily descriptive and oriented toward planning support rather than interactive optimization. Such models highlight the richness of AIS-derived knowledge but also underscore the need for RL frameworks that can move beyond static representations toward adaptive policy learning.

The gaps identified in prior work motivate the development of an RL framework that is both configurable and scalable, capable of operating across multiple origin-destination pairs while integrating real-world environmental data and historical AIS traffic patterns. Table~\ref{tab:comparison} summarizes how our approach compares with existing maritime RL systems, emphasizing its unique combination of large-scale configurability, integration of AIS-derived traffic knowledge, real-time weather modeling, hexagonal spatial representation, and safety mechanisms through action masking.

\section{Methodology}
\label{sec:methodology}

\noindent\textit{Notation.} Table~\ref{tab:symbols} lists the symbols used in the methodology.

\begin{table}[!b]
\centering
\caption{Symbols and Notation}
\label{tab:symbols}
\footnotesize
\setlength{\tabcolsep}{4pt}
\renewcommand{\arraystretch}{1.15}
\begin{tabular}{
    >{\raggedleft\arraybackslash}p{0.25\linewidth}
    >{\raggedright\arraybackslash}p{0.70\linewidth}}
\toprule
\textbf{Symbol} & \textbf{Meaning} \\
\midrule
$\mathcal{S}, \mathcal{A}, \mathcal{G}$ & State, action, and goal spaces \\
$\mathcal{P}, R, \gamma$ & State transition probability, reward function, discount factor \\
$s_t, a_t, r_t$ & State, action, and reward at time $t$ \\
$\pi(a \mid s)$ & Policy (probability of action $a$ in state $s$) \\
$Q^{\pi}(s,a,g)$ & Goal-conditioned action–value function \\
$G=(\mathcal{S},E)$ & Graph of hexagonal cells and transitions \\
$p_{ij}, N_{ij}$ & Transition probability and count from $s_i$ to $s_j$ \\
$W(s,t)$ & Wind vector at cell $s$ and time $t$ \\
$d, \delta$ & Haversine distance and progress-to-goal change \\
$\mathbf{s}_t, \mathbf{s}^{\mathrm{hist}}_t$ & Observation vector and history-augmented state \\
$D(s,g), H(s,g)$ & Shortest-path distance and episode horizon \\
$D_{\mathrm{ref}}$ & Reference task distance (reward scaling) \\
$r_{\mathrm{raw}}, r$ & Raw and scaled rewards \\
$r_{\mathrm{prog}}, r_{\mathrm{freq}}, r_{\mathrm{wind}}$ & Progress, frequency, and wind terms \\
$r_{\mathrm{fuel}}, r_{\mathrm{eta}}, r_{\mathrm{base}}$ & Fuel, time, and step penalties \\
$\text{fuel\_use}, \text{drag}$ & Fuel consumption and drag factor \\
$\mathbf{m}_t(s)$ & Action mask at state $s$ and time $t$ \\
$\mathbf{a}_t = [a^{\mathrm{man}}, a^{\mathrm{spd}}]$ & Composite action: maneuver and speed \\
\bottomrule
\end{tabular}
\end{table}

\subsection{Background}

\subsubsection{Markov Decision Process (MDP)}

Reinforcement learning problems are commonly modeled as a Markov Decision Process (MDP), defined by the tuple $(\mathcal{S}, \mathcal{A}, \mathcal{P}, \mathcal{R}, \gamma)$. At each time step $t$, the agent observes a state $s_t \in \mathcal{S}$, selects an action $a_t \in \mathcal{A}$ according to its policy $\pi(a \mid s)$, receives a reward $r_t = \mathcal{R}(s_t, a_t)$, and transitions to $s_{t+1}$ with probability $\mathcal{P}(s' \mid s, a)$~\cite{sutton2018reinforcement}. The goal of the agent is to learn a policy $\pi$ that maximizes the expected discounted return $\mathbb{E}_\pi\!\left[\sum_{t=0}^{\infty} \gamma^t r_t\right]$.

\subsubsection{Goal-Conditioned Reinforcement Learning (GCRL)}
Standard RL learns a policy tied to a single task and reward, limiting its ability to generalize across objectives. 
GCRL addresses this by conditioning the policy and value function on a goal variable $g \in G$. 
This extends the MDP to a goal-augmented tuple $(S, A, G, P, R, \gamma)$, where $G$ denotes the goal space. 
The resulting policy $\pi(a \mid s, g)$ and value function $Q^{\pi}(s, a, g)$ operate over state--goal pairs, 
allowing agents to reuse learned dynamics across multiple tasks without retraining.

\begin{equation}
\small
Q^{\pi}(s,a,g) = \mathbb{E} \left[ \sum_{t=0}^{\infty} \gamma^t \, r(s_t, a_t, g) \;\middle|\; s_0 = s,\, a_0 = a,\, g \right].
\end{equation}
By conditioning on goals, agents can generalize across tasks and origin--destination pairs with a single unified policy.~\cite{liu2022goal}

\subsubsection{Proximal Policy Optimization (PPO)}
PPO is an on-policy RL algorithm that keeps policy updates stable and sample-efficient. It improves on traditional policy gradients and ensures that policy updates remain within a trusted region~\cite{schulman2017proximal}. At its core, PPO uses a clipped objective to limit policy shifts. It calculates the ratio of new to old policies. The Objective is:
\begin{equation}
L^{\mathrm{CLIP}} = \hat{\mathbb{E}}_t \left[ \min \left( r_t \hat{A}_t, \, \text{clip}(r_t, 1-\epsilon, 1+\epsilon) \hat{A}_t \right) \right]
\end{equation}

where $\hat{A}_t$ estimates advantages, and $\epsilon$ (around 0.1--0.2) clips ratios to stay close to $1$.

\subsubsection{Random Network Distillation (RND) for Exploration}
RND drives exploration with intrinsic rewards based on prediction-based curiosity, making it highly effective for discovering routes in sparse-reward maritime environments~\cite{burda2018exploration}. RND uses a fixed, randomly initialized target neural network $f$ and another network, known as the predictor, $\hat{f}$, trained to approximate it. The intrinsic reward is formally defined as:
\begin{equation}
r_t^{\mathrm{intrinsic}} = \lVert \hat{f}(s_t) - f(s_t) \rVert_2^2.
\end{equation}

\noindent
This term augments the environmental reward with weight $\beta$:
\begin{equation}
r_t^{\mathrm{total}} = r_t^{\mathrm{extrinsic}} + \beta \cdot r_t^{\mathrm{intrinsic}}.
\end{equation}

By rewarding the agent for visiting states that yield high prediction errors, RND promotes curiosity-driven behavior. This helps the agent to escape local optima and explore underrepresented regions of the state space, leading to more diverse trajectories and improved chances of discovering globally optimal navigation strategies given the state space.

\subsubsection{Haversine Distance}

It calculates the great-circle distance $d$ between two points on a sphere, given their latitudes ($\phi$) and longitudes ($\lambda$), accounting for Earth's curvature~\cite{rahayu2024optimizing}:
\begin{multline}
d = 2R \arcsin \Bigg( 
\sqrt{ \sin^2\!\left(\tfrac{\Delta \phi}{2}\right) 
+ \cos(\phi_1)\cos(\phi_2) } \; \cdot \\
\sqrt{ \sin^2\!\left(\tfrac{\Delta \lambda}{2}\right) } 
\Bigg)
\end{multline}
where $\Delta\phi$ and $\Delta\lambda$ are the differences in latitude and longitude between the points, $\phi_1$ and $\phi_2$ are their respective latitudes, and $R$ is Earth's mean radius ($\approx 6{,}371$ km).

\subsubsection{Action Masking}
\begin{figure}[h]
    \centering
    \includegraphics[width=0.45\linewidth]{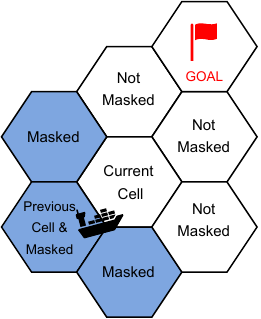}
    \caption{Illustration of action masking. From the current cell, invalid moves such as land or backtracking are masked in blue, while valid navigable directions remain available to the agent.}
    \label{fig:action_masking}
\end{figure}

Action masking is a reinforcement learning technique that restricts the policy to select only valid or state-relevant actions by applying a binary or continuous mask to the action space during sampling and/or optimization.~\cite{huang2020closer}. This technique prevents the agent from choosing impossible moves, such as moving into land-based cells or returning to its immediate previous cell. As illustrated in Fig.~\ref{fig:action_masking}, the set of valid actions is dynamically updated based on the agent's current state.

The process applies a binary mask $\mathbf{m}_t(s)$ to the action space, assigning $1$ to valid actions and $0$ to invalid ones:
\begin{equation}
\mathbf{m}_t(s)[a] = \begin{cases}
1 & \text{if action } a \text{ is valid in state } s, \\
0 & \text{if action } a \text{ is invalid in state } s.
\end{cases}
\end{equation}

\noindent
The policy probabilities are re-normalized over valid actions:
\begin{equation}
\pi_{\mathrm{masked}}(a \mid s) = \frac{\pi(a \mid s) \cdot \mathbf{m}_t(s)[a]}{\sum_{a' \in \mathcal{A}} \pi(a' \mid s) \cdot \mathbf{m}_t(s)[a']}.
\end{equation}

\subsubsection{Hexagonal Geospatial Indexing (H3)}
\begin{figure}[ht]
    \centering
    \includegraphics[width=0.25\columnwidth]{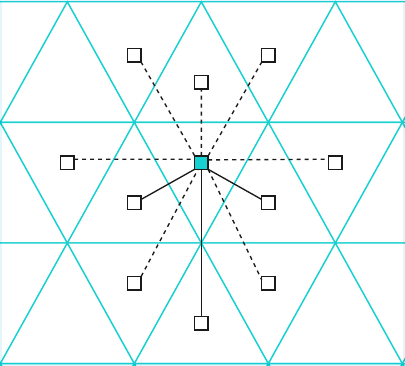}\hspace{0.9em}
    \includegraphics[width=0.25\columnwidth, height=2cm]{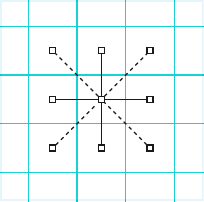}\hspace{0.9em}
    \includegraphics[width=0.25\columnwidth]{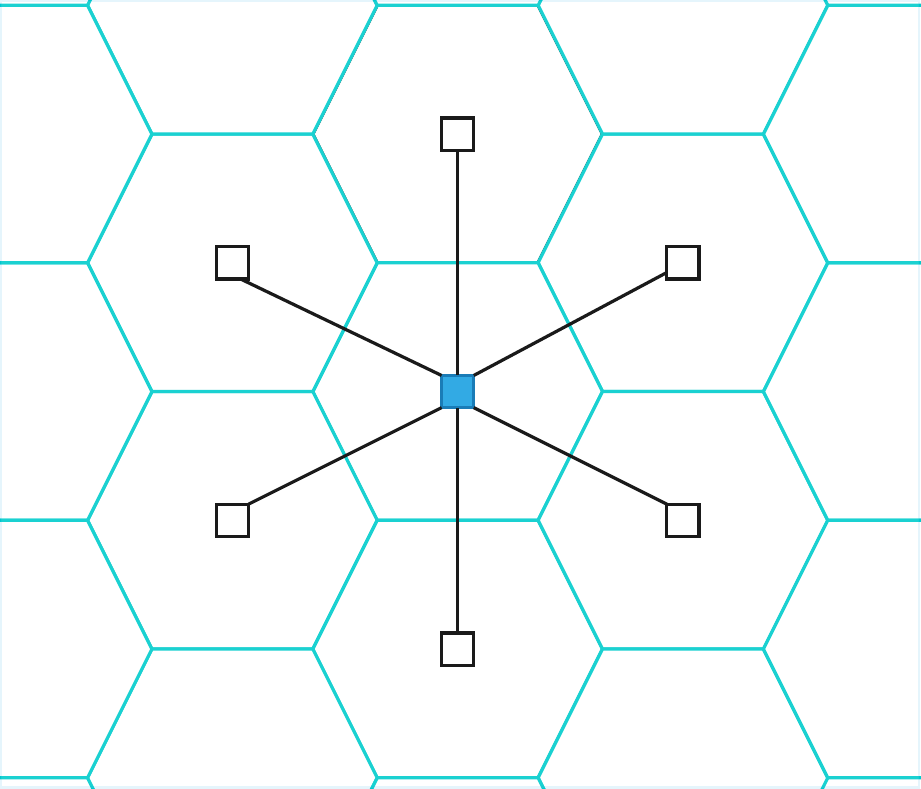}%
    \caption{Distances from a triangle to its neighbors (left), a square to its neighbors (center), and a hexagon to its neighbors (right).}
    \label{fig:H3}
\end{figure}

H3, Uber's hierarchical hexagonal geospatial indexing system, partitions the Earth into a multi‑resolution hexagonal grid~\cite{Brodsky_H3_Uber}. Its key advantage over square grids is the ``one‑distance rule," where all neighbors of a hexagon lie at comparable step distances. As illustrated in Fig.~\ref{fig:H3}, this uniformity removes the diagonal-versus-edge ambiguity present in square lattices. In turn, it simplifies routing and spatial analysis by yielding consistent step costs, which is central to our simulation setup.

\subsection{Data Processing and Spatial Discretization}
\label{subsec:data_processing}

This section corresponds to the AISdb block at the left of Fig.~\ref{fig:MariNav_Arch_diag}, which standardizes trajectories and discretizes space prior to graph construction.

\subsubsection{AIS Data Pre-Processing}
We use satellite-based AIS records spanning 2013--2024, sourced from Kpler. This dataset includes over 10 million records from the Gulf of St. Lawrence, each with navigational attributes such as timestamps, latitude/longitude, Speed over Ground, and Course over Ground. This region was selected for its dense maritime traffic and environmental variability. Raw AIS trajectories have irregular temporal intervals (1 to 60+ minutes). To create standardized trajectories for modeling, we used AISdb~\cite{spadon2024maritime} (an open-source platform for integrating vessel tracking and environmental data) to resample positions at uniform 1-minute intervals via linear interpolation. This produces a temporally consistent trajectory $\tilde{\tau}_v$ for each vessel $v$:
\begin{equation}
\tilde{\tau}_v = \{(x_{t'}, y_{t'})\}_{t' \in T'}.
\end{equation}
where $x_{t'}$ and $y_{t'}$ denote longitude and latitude at resampled time $t'$ and $T'$ represents the set of resampled timestamps~\cite{spadon2025learning}.

\subsubsection{Hexagonal Spatial Discretization}

We discretize the Gulf of St. Lawrence (bounding box [-74.9540, 44.9336, -54.6965, 52.2518]) into a hexagonal grid with H3 at resolution 6. Using a public coastline boundary file, we clip the grid to the study area so it covers the Gulf and excludes land and out-of-scope waters. This procedure yields 8,687 hexagons, each about 36 km$^2$. We then map each resampled position $(x_{t'},y_{t'})$ to a cell $s_i \in S$, where $S$ is the set of hexagons retained after clipping.

\subsection{Markovian Graph Construction}
\label{subsec:markovian_graph}

\begin{figure*}[ht]
    \centering
    \includegraphics[width=\linewidth]{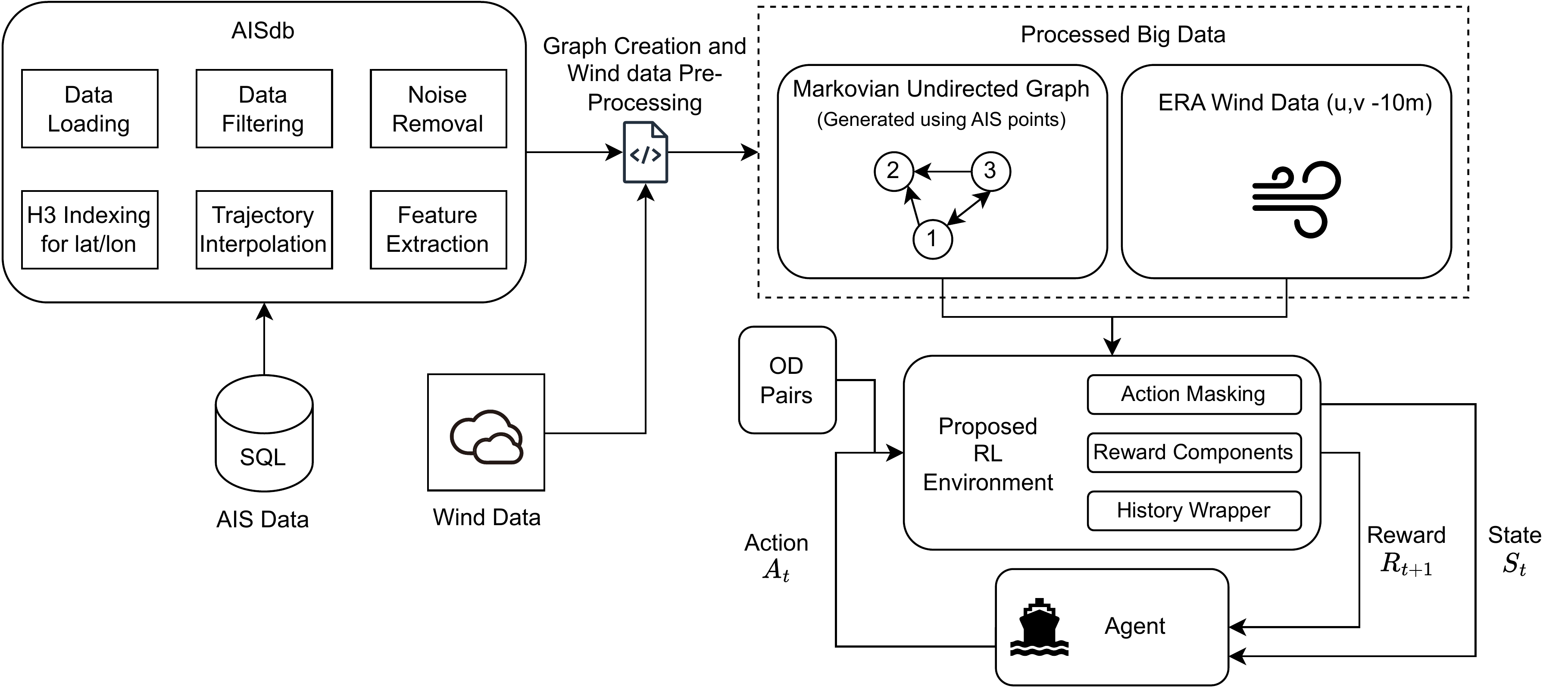}%
    \vspace{-5pt}
    \caption{Framework overview. AISdb denoises AIS trajectories, which are converted into a Markovian traffic graph. ERA5 wind fields are incorporated separately, and together with the graph and weather conditions, into the RL environment. The environment applies action masking, reward shaping, and optional history wrapping to support multi-origin–destination tasks.}
    \label{fig:MariNav_Arch_diag}
\end{figure*}

We model historical traffic flow as a weighted undirected graph where nodes represent hexagonal cells. 
Following~\cite{spadon2025learning}, vessel trajectories are assumed to follow a  first-order Markov property, where the next state depends only on the current state 
(Fig.~\ref{fig:MariNav_Arch_diag}, \textit{Processed Big Data} block). Formally, let $X_t \in \mathcal{S}$ be the hexagonal state at discrete timestep $t$. Vessel mobility is then modeled as:
\begin{multline}
P(X_{t+1} = s_j \mid X_t = s_i, X_{t-1}, \ldots, X_0) \\
= P(X_{t+1} = s_j \mid X_t = s_i).
\end{multline}

For any adjacent cells 
$s_i$ and $s_j$, there is a single edge connecting them, which carries symmetric attributes. 
The edge weight represents the transition probability, defined as the average of normalized 
transition frequencies in both directions. Each edge also stores the mean observed vessel 
speed along that edge, and the traversal cost is computed as the great-circle distance divided by edge speed, 
plus a reliability penalty scaled by a factor $\lambda \geq 0$, and will be used to evaluate graph-based baseline paths in our experiments.

\subsection{Wind Dynamics Modeling}
\label{subsec:wind_map}

The second data source is the ERA5 global atmospheric reanalysis dataset~\cite{hersbach2020era5}. We use ERA5 reanalysis (hourly) from the Copernicus Climate Data Store. Wind fields are integrated into our spatial grid as a temporal map:
\begin{equation}
W: \mathcal{S} \times T \rightarrow \mathbb{R}^2,
\end{equation}
which maps each cell $s_i \in \mathcal{S}$ and time $t \in T$ to a wind vector $(u_{w,10m}, v_{w,10m})$, where $u_{w,10m}$ and $v_{w,10m}$ are the $10$-m east--west and north--south components.  

\subsection{Reinforcement Learning Environment Design}
\label{subsec:rl_environment}

The proposed RL framework, with environmental inputs and agent interactions, is shown in Fig.~\ref{fig:MariNav_Arch_diag}.

\subsubsection{State Space}
The state representation specifies what the agent knows at each step, including current environmental conditions and the navigation goal. In our formulation, the policy is conditioned on the vessel's geospatial location, its selected speed, the local wind direction, and the fixed navigation goal defined by start and destination coordinates. The resulting observation vector at time $t$ is expressed as an 8-dimensional feature-normalized vector:
\begin{align}
\mathbf{s}_t = [
    & \text{lat}^{(t)}_{\mathrm{norm}}, \; \text{lon}^{(t)}_{\mathrm{norm}}, \nonumber \\
    & \text{speed}^{(t)}_{\mathrm{norm}}, \; \text{wind\_dir}^{(t)}_{\mathrm{norm}}, \nonumber \\
    & \text{start\_lat}_{\mathrm{norm}}, \; \text{start\_lon}_{\mathrm{norm}}, \nonumber \\
    & \text{goal\_lat}_{\mathrm{norm}}, \; \text{goal\_lon}_{\mathrm{norm}}
].
\end{align}

To ensure consistency across heterogeneous variables, each feature undergoes normalization tailored to its physical domain. Linear min--max scaling is applied to bounded positional variables such as latitude and longitude:
\begin{equation}
x_{\mathrm{norm}} = \frac{x - x_{\min}}{x_{\max} - x_{\min}},
\end{equation}
While positively skewed variables such as speed are log-transformed, preventing dominance of extreme values:
\begin{equation}
y_{\mathrm{norm}} = 
\frac{\log_{10}(y + \epsilon) - \log_{10}(y_{\min} + \epsilon)}%
     {\log_{10}(y_{\max} + \epsilon) - \log_{10}(y_{\min} + \epsilon)},
\end{equation}
with $\epsilon=10^{-5}$ to avoid singularities at zero. Wind direction is normalized by mapping its range $[-\pi,\pi]$ onto $[0,1]$, yielding a consistent scale while eliminating the discontinuity at the angular boundary:
\begin{equation}
\text{wind\_dir}_{\mathrm{norm}} = \frac{\text{wind\_dir} + \pi}{2\pi}.
\end{equation}

The extrema $(x_{\min},x_{\max},y_{\min},y_{\max})$ are computed once per episode based on all H3 nodes in the navigation graph, ensuring environment-specific calibration. Start and goal coordinates are normalized analogously to the agent's position, using the cell center values of their respective H3 indices.  

\subsubsection{History-Augmented State}
\label{subsec:history_augmented_state}
To capture temporal dependencies, we extend the base observation with a configurable history of dynamic features. Let the base state decompose as:
\begin{equation}
\mathbf{s}_t = [\mathbf{z}_t, \mathbf{g}],
\end{equation}
where:
\begin{itemize}
    \item $\mathbf{z}_t \in \mathbb{R}^{d_z}$ represents the dynamic variables that evolve over time (e.g., position, speed, wind).
    \item $\mathbf{g} \in \mathbb{R}^{d_g}$ encodes the static navigation information, namely the start and goal coordinates.
\end{itemize}
In our setup, the total feature dimension is $D=8$. Of these, four correspond to the start--goal specification ($d_g=4$), leaving four dynamic variables ($d_z = D - d_g = 4$).

Given a user-defined history length $H \in \mathbb{N}$ (default $H=8$), the history-augmented state concatenates $H$ recent dynamic observations with the static goal descriptor:
\begin{equation}
\mathbf{s}^{\mathrm{hist}}_t = \big[ \mathbf{z}_{t-H+1}, \dots, \mathbf{z}_{t}, \mathbf{g} \big] \in \mathbb{R}^{H d_z + d_g}.
\end{equation}

At the beginning of an episode, when fewer than $H$ observations are available ($t < H-1$), the missing entries are padded by replicating the initial dynamic vector $\mathbf{z}_0$. The static goal descriptor $\mathbf{g}$ is appended once, ensuring a consistent state dimension independent of history length. This augmentation equips the agent with short-term trajectory context while preserving the invariant navigation objective.

\subsubsection{Episode Horizon and Termination}
The temporal extent of each episode is determined by the shortest-path distance between a start state $s_0$ and a goal $g$, denoted $D(s_0,g)$. This distance corresponds to the number of hexagonal cell transitions on the navigation graph, as described in Section~\ref{subsec:markovian_graph}. Distances are computed using Dijkstra's algorithm. An episode terminates successfully if the agent reaches the goal, and it is truncated if the horizon $H(s,g) := 5 \cdot D(s,g)$ is reached beforehand.

\subsubsection{Reward and GCRL Scaling}
\label{sec:reward}
The reward function combines several components to encourage efficient, safe, and realistic navigation. The raw reward is defined as:
\begin{equation}
r_{\mathrm{raw}} = r_{\mathrm{prog}} + r_{\mathrm{freq}} + r_{\mathrm{wind}} + r_{\mathrm{fuel}} + r_{\mathrm{eta}} + r_{\mathrm{base}},
\end{equation}
where each term provides a distinct incentive:
\begin{align}
r_{\mathrm{prog}} &= 2 \cdot \delta, \\
r_{\mathrm{freq}} &= \mathrm{clip}\!\left( \frac{\log(1 + \text{edge\_weight})}{5}, 0, 0.5 \right), \\
r_{\mathrm{wind}} &= 
\begin{cases}
-1 & \text{if } \text{wind\_speed} > 10 \ \text{m/s}, \\
0  & \text{otherwise},
\end{cases} \\
r_{\mathrm{fuel}} &= -0.001 \cdot \text{fuel\_use}, \\
r_{\mathrm{eta}}  &= -0.001 \cdot \text{travel\_time}, \\
r_{\mathrm{base}} &= -1.
\end{align}

Here, $r_{\mathrm{prog}}$ rewards progress toward the goal, where
\[
\delta = D(s_{t}, g) - D(s_{t+1}, g),
\]
represents the difference in shortest-path distances through the graph. The term $r_{\mathrm{freq}}$ guides the agent toward frequently traveled routes derived from the edge weights of the Markovian navigation graph. Although this bias can be suboptimal, as less-visited states may yield better solutions, it does not dominate the reward function. Instead, it acts as a soft preference, indicating that cells traversed by many vessels are likely safe for the agent to enter. The remaining terms penalize exposure to strong winds, fuel consumption, and prolonged travel time, while $r_{\mathrm{base}}$ applies a constant per-step penalty, promoting shorter paths.

Fuel consumption is modeled using a cubic-law relation adapted from marine engineering studies~\cite{Kim2020ISO15016Fuel, Zhang2020AirResistanceContainerShip, Wang2019ObliqueWindModel}:
\begin{equation}
\label{eq:fuel_consumption}
\text{fuel\_use} = 0.05 \times \text{speed}^3 \times \text{drag} + 0.02 \times \text{wind\_speed}
\end{equation}
where
\begin{equation}
\text{drag} = 1 + 0.5 \cdot \big(1 - \cos(\text{heading} - \text{wind\_dir})\big)
\end{equation}
captures additional resistance due to headwinds~\cite{Zhang2021SpeedOptimizationFleet}. To maintain numerical stability, the fuel penalty is scaled by $-0.001$, preventing the cubic law from dominating the overall reward.

To make rewards comparable across GCRL, we rescale by task length relative to a fixed reference:
\begin{equation}
r = \frac{r_{\mathrm{raw}}}{D(s,g)} \cdot D_{\mathrm{ref}}, \quad
D_{\mathrm{ref}} = D(s_{\mathrm{ref}}, g_{\mathrm{ref}}),
\label{eqn:27}
\end{equation}

where $(s_{\mathrm{ref}}, g_{\mathrm{ref}})$ is a fixed reference pair. This ensures consistency, aligning magnitudes across tasks of different lengths.

Lastly, invalid actions (e.g., moving to a non-neighbor in an unmasked setting) incur an unscaled terminal penalty of $r = -1900$, overriding (\ref{eqn:27}), ensuring they are strongly discouraged regardless of route length.

\subsubsection{Periodic Weighted Sampling of Start--Goal Tasks}
To ensure balanced coverage of training tasks, we adopt a periodic weighted sampling scheme. Let $\mathcal{T} = \{(s_i,g_i)\}_{i=1}^N$ denote the set of training tasks and $c_i$ the number of completed episodes for task $i$. Training alternates between two modes: uniform sampling, applied by default, and coverage-oriented sampling, activated periodically for brief windows. After every per-environment threshold of $S_{\mathrm{th}} = 62{,}500$ steps, a prioritized window of length $W=6{,}250$ steps is triggered, which under parallel execution approximates a global ``1M/10k" schedule. During the prioritized window, each task $i$ is assigned a weight
\[
w_i = \max_{j=1,\dots,N} c_j - c_i + 1,
\]
where $c_j$ is the number of completed episodes for task $j$ and the maximum is taken over all tasks. The sampling probability for task $i$ is then
\[
p_i = \frac{w_i}{\sum_{k=1}^N w_k}.
\]
Outside the window, uniform sampling is restored.

This mechanism prevents rare or complex tasks from being neglected, ensures balanced task coverage, and stabilizes learning by periodically catching up on under-trained tasks.

\subsubsection{Action Space}
The action space is multi-discrete: the agent selects a maneuver and a speed at each time step. The action at time $t$ is given as:
\begin{equation}
\mathbf{a}_t = \big[a^{\mathrm{maneuver}}_t, \; a^{\mathrm{speed}}_t\big],
\end{equation}
where $a^{\mathrm{maneuver}}_t \in \mathcal{A}_{\mathrm{man}}$ defines the next cell transition and $a^{\mathrm{speed}}_t \in \mathcal{A}_{\mathrm{spd}}$ specifies the vessel's cruising speed.

The maneuver component $\mathcal{A}_{\mathrm{man}}$ consists of transitions to neighboring hexagonal cells in the H3 navigation graph. Each hexagon has six geometric neighbors; however, the agent is restricted to at most five options since direct backtracking to the previous cell is prohibited. This restriction enforces forward progress and prevents oscillatory behavior.  

The speed component $\mathcal{A}_{\mathrm{spd}}$ is discretized into five operational levels ranging from $8$ to $22$ knots:
\begin{equation}
\mathcal{A}_{\mathrm{spd}} = \{8, 11, 14, 18, 22\}\ \text{knots}.
\end{equation}
This range reflects realistic operating conditions, as commercial vessels typically navigate within this interval to balance fuel efficiency and mechanical reliability~\cite{latinopoulos2025marine}. Discretization also reduces variance in policy updates by constraining the continuous control problem to a manageable finite set.  

To ensure safety and efficiency, we apply a dynamic action mask derived from the weighted, undirected navigation graph. The mask enforces three constraints: (i) the agent may only transition to valid successor cells defined in the graph, (ii) the agent cannot return to the previous cell, and (iii) at the start of an episode, if the initial cell has six neighbors, the transition with the lowest edge weight is masked to steer exploration away from the least promising neighbor.

%%%=====================================================================%%%
\section{Experimental Evaluation}
\label{sec:evaluation}

\subsection{Dataset}
\label{sec:dataset}
\begin{figure*}[htbp]
    \centering
    \begin{subfigure}[b]{0.32\linewidth}
        \centering
        \includegraphics[width=\linewidth]{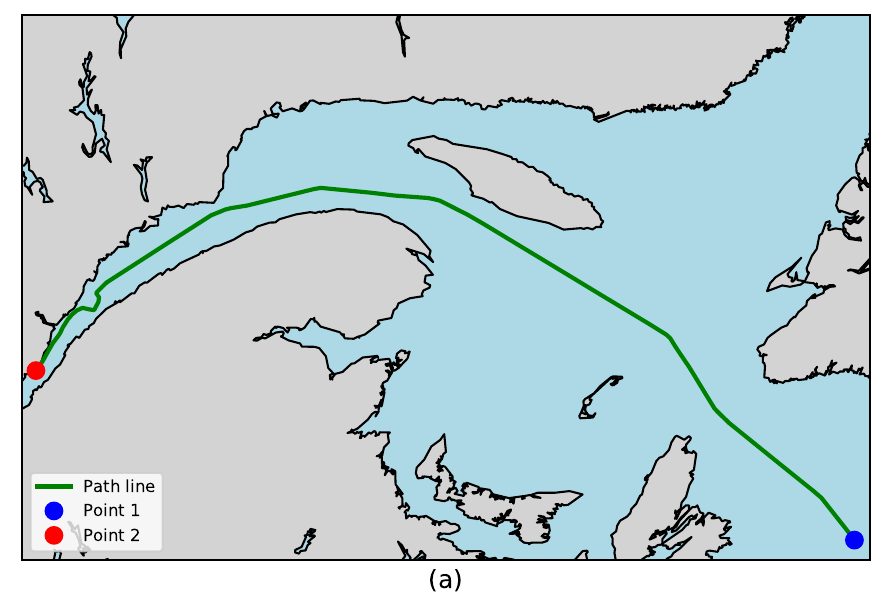}
        \label{fig:path1}
    \end{subfigure}
    \hfill
    \begin{subfigure}[b]{0.32\linewidth}
        \centering
        \includegraphics[width=\linewidth]{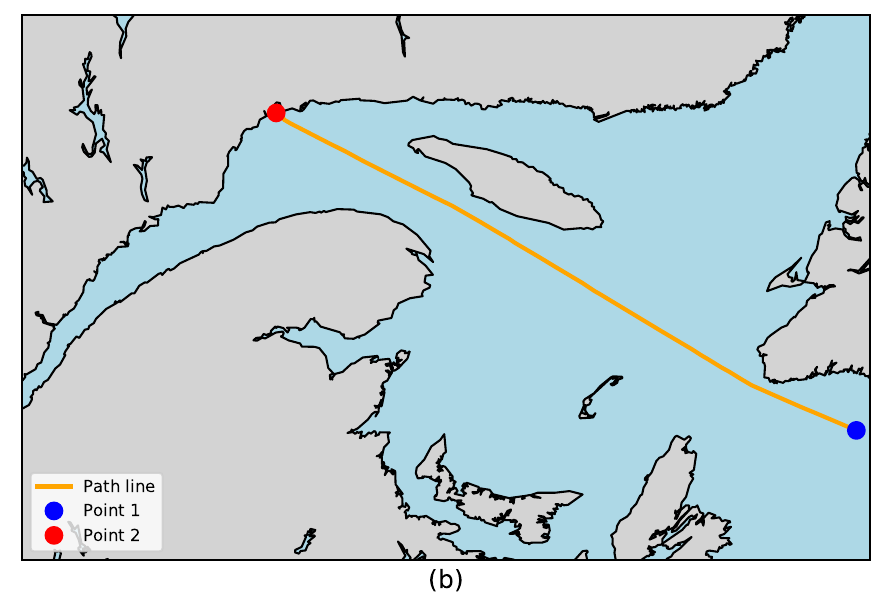}
        \label{fig:path2}
    \end{subfigure}
    \hfill
    \begin{subfigure}[b]{0.32\linewidth}
        \centering
        \includegraphics[width=\linewidth]{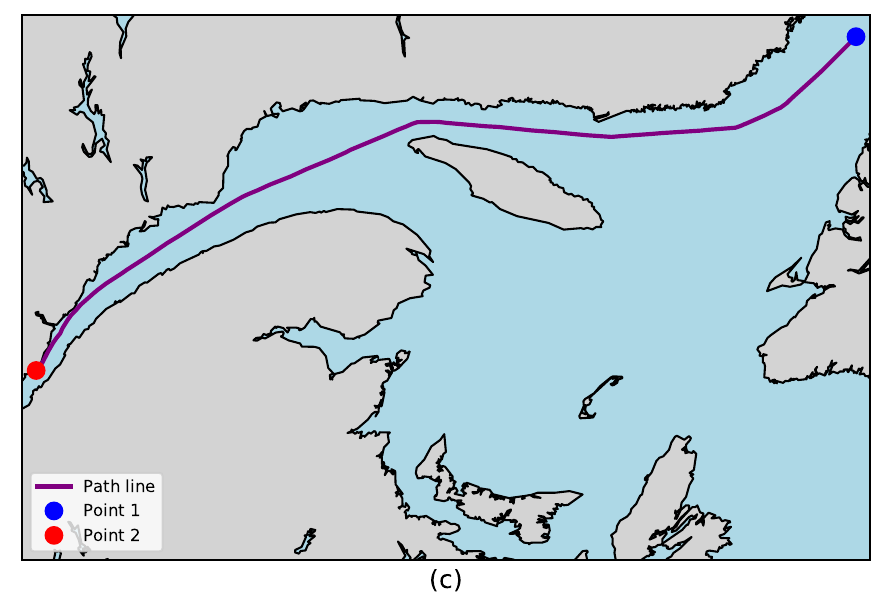}
        \label{fig:path3}
    \end{subfigure}
    \vspace{-7pt}
    \caption{Baseline vessel routes in the Gulf of St. Lawrence. The three representative tracks are: (a) Path 1: entering the Gulf toward Ontario, (b) Path 2: near-straight toward Sept-Îles, and (c) Path 3: curved from the Strait of Belle Isle toward Ontario.}
    \label{fig:paths}
\end{figure*}

We train and evaluate our framework on AIS trajectories from the Gulf of St. Lawrence seaways, focusing on tanker and cargo vessel trips. To construct the navigation environment, we build a year-long traffic graph from all 2024 AIS records of cargo and tanker vessels, capturing corridor-level connectivity patterns. Wind dynamics are incorporated from ERA5 hourly 10-meter fields for August 2024. During training, each episode randomly samples a day from this month, ensuring that the agent is exposed to the full range of August weather conditions across episodes. For experiments, six representative origin--destination corridors are defined. Fig.~\ref{fig:paths} shows three representative tracks, traversed in both directions, covering all six training tasks.

\subsection{Reproducibility}
\label{subsec:reproducibility}

Our results were condensed in a tool for easy reproducibility named \textit{MariNav}\footnote{https://github.com/Vaishnav2804/MariNav}, which is a Gymnasium-compatible environment for RL in maritime navigation. All experiments were conducted using Stable-Baselines3 (SB3)~\cite{stable-baselines3} and SB3-Contrib. To speed up data collection, each run utilized a vectorized setup with 16 parallel environments running in separate processes, allowing for the simultaneous gathering of multiple rollouts. For stable training across runs, we applied environment-level normalization, which standardizes rewards while leaving observations unchanged, thereby ensuring comparable learning dynamics across experiments. The full set of training hyperparameters is summarized in Table~\ref{tab:rl_hyperparameters}.

\begin{table}[!htb]
  \centering
  \caption{Hyperparameters used for the RL agent.}
  \label{tab:rl_hyperparameters}
  \begin{tabular}{l l}
    \toprule
    Label & Value \\
    \midrule
    Policy architecture & Two-layer MLP (shared actor--critic) \\
    Learning rate & Linear decay from \(7\times10^{-4}\) to \(1\times10^{-5}\) \\
    Entropy coefficient & \(0.01\) \\
    Rollout length (\(n_{\text{steps}}\)) & \(1024\) per worker \\
    Mini-batch size & \(64\) (with gradient accumulation) \\
    Epochs per update & \(10\) \\
    Parallel environments & \(16\) \\
    Discount factor (\(\gamma\)) & \(0.99\) \\
    Device & CPU; deterministic seeding across frameworks \\
    Random seeds & 3, 31, 42 \\
    \bottomrule
  \end{tabular}
\end{table}

\subsection{Results \& Discussions}
\label{sec:results}

\begin{figure*}[t]
    \centering
    \begin{subfigure}{0.48\linewidth}
        \includegraphics[width=\linewidth]{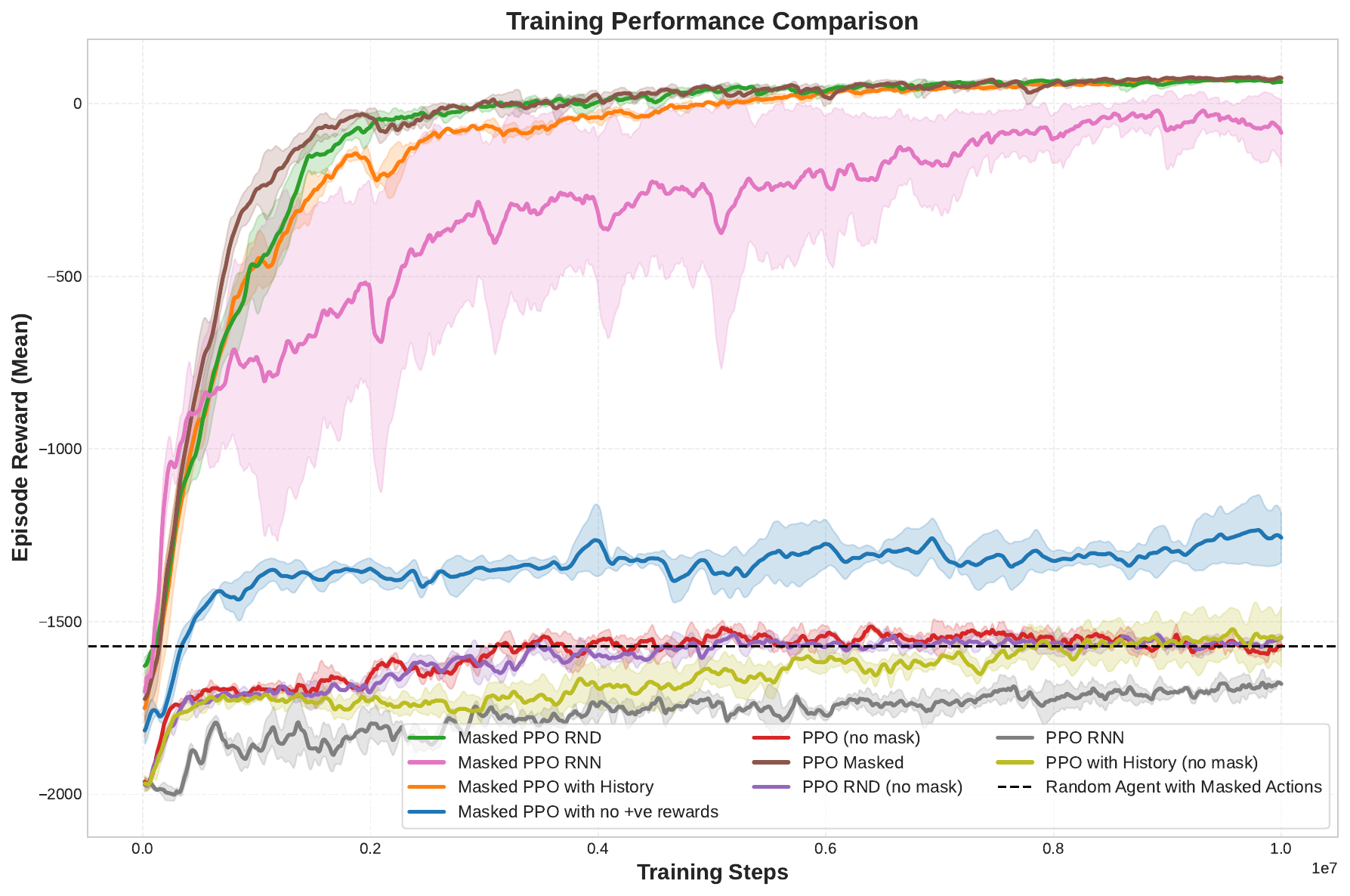}
        \caption{Full reward setting}
        \label{fig:reward_comparison}
    \end{subfigure}
    \hfill
    \begin{subfigure}{0.48\linewidth}
        \includegraphics[width=\linewidth]{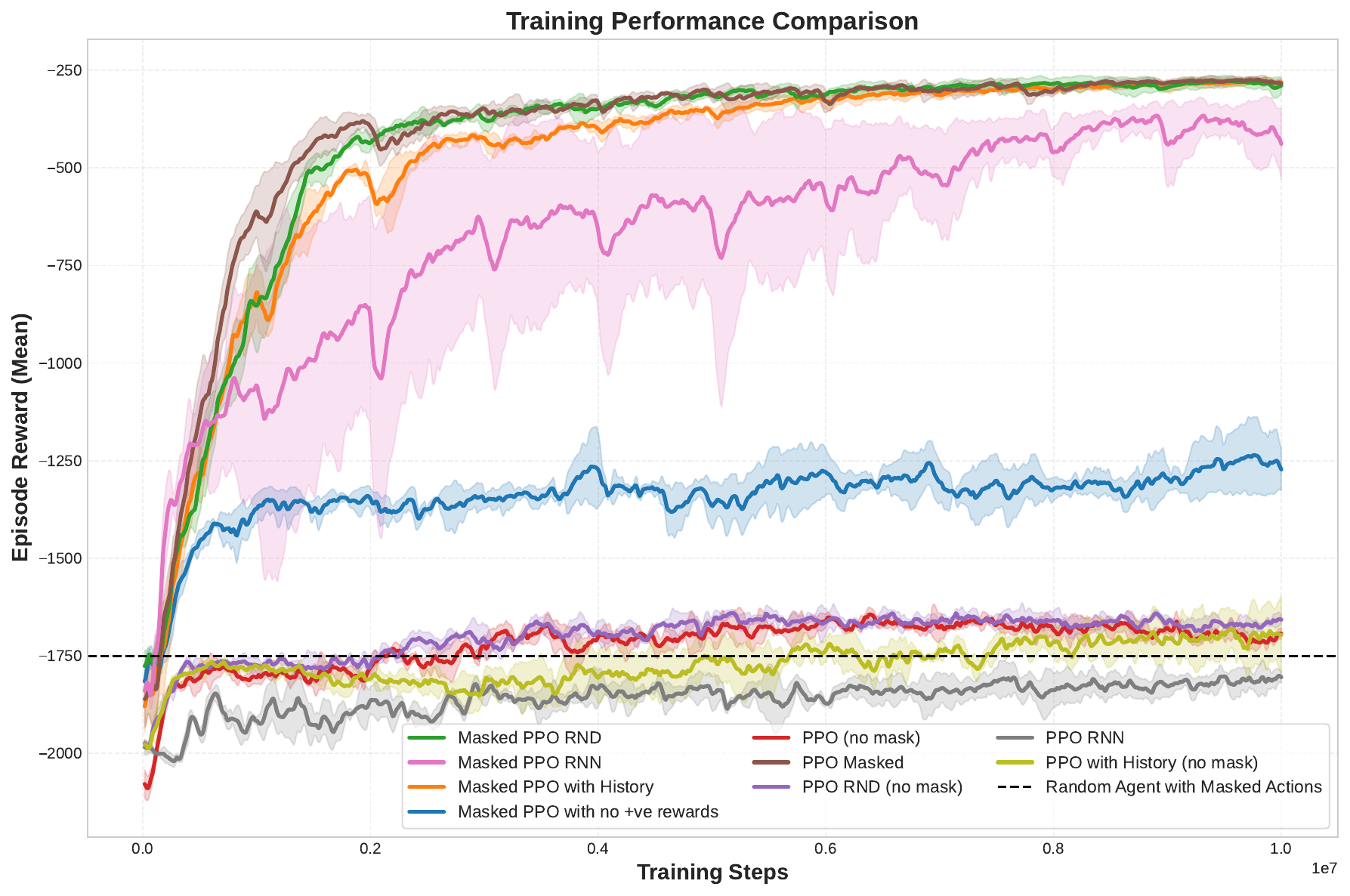}
        \caption{Penalty-only setting}
        \label{fig:reward_comparison_neg}
    \end{subfigure}
    \caption{Mean episode return versus training steps for each policy variant. 
    Solid lines show seed-averaged performance; shaded bands indicate $\pm1$ s.d. across three seeds.}
    \label{fig:reward_comparison_combined}
\end{figure*}

As an initial step in the evaluation, we tested nine policy variants to assess how specific design choices influence learning outcomes in maritime navigation. In particular, we examined the role of action masking in enforcing feasibility constraints, the effect of including observation history—both via history-augmented states and through recurrent networks (RNNs)—on temporal consistency, and the contribution of intrinsic exploration through RND. Table~\ref{tab:policy_variants} reports the mean return over the final 100 episodes for all variants, averaged across three seeds, along with standard deviations, while Fig.~\ref{fig:reward_comparison} complements these results by illustrating the learning curves and variance over ten million steps. For completeness, because the true objective of the agent is to minimize the penalty terms as discussed in Section~\ref{sec:reward}, we report returns under the penalty-only formulation of rewards, as well (see Fig.~\ref{fig:reward_comparison_neg}).

For clarity in the subsequent discussion, we refer to the different policy variants as follows. \textit{Masked PPO} is the PPO agent with the action-masking scheme described above. Variants with \textit{+History} use the history-augmented state, while \textit{+RND} and \textit{+LSTM} indicate the addition of intrinsic exploration or a recurrent network, respectively. \textit{PPO (no mask)} refers to the standard PPO agent without action masking.

%Add this to above%
% The environment combines AIS-derived vessel mobility graphs with ERA5 wind dynamics on a hexagonal grid, providing a structured and data-driven setting for training and evaluation. \textit{MariNav} also includes an extension, \texttt{MariNavWithHistory}, which augments observations with stacked time histories to support recurrent policy architectures. This enables experiments that account for short-term temporal dependencies while retaining a consistent interface with standard RL algorithms.  
%  

\begin{table}[h]
\centering
\caption{Mean return (last 100 episodes; grouped by policy variants; best variant first).}
\begin{tabular}{lr}
\toprule
Approach & Mean Return ± Std. Dev. \\
\midrule
\textbf{Masked PPO} & $68.03 \pm 2.45$ \\ 
PPO (no mask) & $-1556.56 \pm 10.06$ \\
\cmidrule(lr){1-2}
Masked PPO + History & $60.94 \pm 6.48$ \\ 
PPO + History (no mask) & $-1473.87 \pm 97.44$ \\
\cmidrule(lr){1-2}
Masked PPO + RND & $-6.97 \pm 100.92$ \\ 
PPO + RND (no mask) & $-1548.69 \pm 14.67$ \\
\cmidrule(lr){1-2}
Masked PPO + LSTM & $-40.47 \pm 60.22$ \\ 
PPO + LSTM & $-1706.98 \pm 14.00$ \\
\cmidrule(lr){1-2}
Masked PPO with no +ve rewards & $-1295.14 \pm 18.88$ \\ 
\bottomrule
\end{tabular}
\label{tab:policy_variants}
\end{table}

\subsubsection{Effect of Graph-Based Shaping Rewards}
To evaluate the necessity of graph-based positive shaping rewards, we trained agents in a penalty-only setting. In this baseline, the learning signal consisted exclusively of penalty terms, including wind exposure, fuel consumption, ETA violations, per-step costs, and invalid-move penalties. To enable a fair comparison with agents trained under the full reward scheme, we report both agents’ performance after removing the penalty terms from evaluation. The results show that without the graph-derived shaping rewards the agent makes little meaningful progress. Consequently, all subsequent experiments employ the mixed reward formulation (penalties + graph-based shaping rewards).

\vspace{0.cm}

% \subsubsection{Impact of Action Masking, Temporal Context, Exploration, and Recurrence}

\subsubsection{Action Masking Is Critical}
The most visible difference arises between masked and unmasked PPO. With action masking, agents achieve consistently high positive returns, whereas unmasked agents fail catastrophically due to frequent selection of invalid actions, which prematurely terminate episodes. Even a randomly initialized policy with masking can settle at a stable return level that unmasked agents only approach after extensive training, illustrating masking's role in feasibility and exploration control. Overall, masking improves learning efficiency and stability by excluding impossible actions and narrowing the effective search space during policy optimization.

\subsubsection{Effect of Short-Term Observation History}
We evaluate a history-augmented state (Section~\ref{subsec:history_augmented_state}) by stacking the most recent  $8$ observations, and comparing masked and unmasked PPO under identical settings. Incorporating this history did not materially change outcomes: the masked variant remained stable, and the unmasked variant remained unstable. The nearly identical averages suggest that the environment is already close to satisfying the Markov property under the chosen state representation.

\subsubsection{Effect of Intrinsic Exploration (RND)}
Adding an RND exploration bonus provides limited benefit. It encourages broader exploration early in training, but the effect diminishes as the predictor network adapts. The learning converges to a suboptimal performance, indicating potential interference from the intrinsic rewards.

\subsubsection{Effect of Recurrent Architectures (RNNs)}
Augmenting Masked PPO with a recurrent network (\textit{Masked PPO RNN}) underperforms relative to both standard Masked PPO and the short-history variant. In this near-Markovian environment, the additional complexity of RNNs appears to hinder learning, producing high variance across seeds. These results suggest that short histories or feedforward policies provide more stable learning dynamics in this setting.

Overall, these findings reveal a clear hierarchy: action masking is essential for feasible policies, short histories stabilize training, and exploration bonuses have only minor impact, while recurrent architectures offer no additional benefit in this setting.

\noindent
% \begin{figure}
% \includegraphics[width=\linewidth]{images/reward_comparison.pdf}
% \caption{Mean episode return versus training steps for each policy variant in the \textbf{full reward setting}. Solid lines show seed-averaged performance; shaded bands indicate ±1 s.d. across three seeds.}
% \label{fig:reward_comparison}
% \end{figure}

% \begin{figure}
% \includegraphics[width=\linewidth]{images/reward_comparison_neg.pdf}
% \caption{Mean episode return versus training steps for each policy variant in the \textbf{penalty-only setting}. Solid lines show seed-averaged performance; shaded bands indicate ±1 s.d. across three seeds.}
% \label{fig:reward_comparison_neg}
% \end{figure}

\begin{figure}
    \centering
    \includegraphics[width=\linewidth]{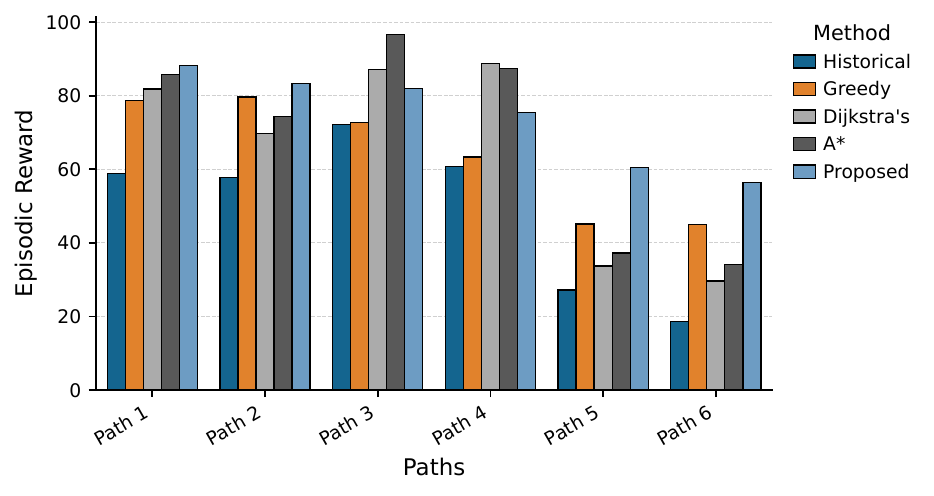}
    \caption{Episodic reward comparison over six routes between the RL-based routing method and baseline approaches. Paths 1-6 are based on the three pairs of points, as discussed in Section~\ref{sec:dataset}}
    \label{fig:baselines}
\end{figure}
\subsubsection{Comparison to Baselines}

% \begin{table}[!b]
% \centering
% \caption{Performance comparison.}
% \label{tab:baseline_results}
% \begin{tabular}{lccc}
% \toprule
% \textbf{Method} & \textbf{Mean} & \textbf{Fuel Use} & \textbf{ETA Deviation} \\
% \midrule
% A* over distance & -- & -- & -- \\
% Greedy over distance & -- & -- & -- \\
% Markovian Traffic Graph Policy & -- & -- & -- \\
% Historical Route & -- & -- & -- \\
% ATRN~\cite{zhang2025adaptive} & -- & -- & -- \\
% Weather Routing~\cite{latinopoulos2025marine} & -- & -- & -- \\
% DQN Maritime~\cite{alam2023ai} & -- & -- & -- \\
% \midrule
% \textbf{Proposed} & -- & -- & -- \\
% \bottomrule
% \end{tabular}
% \end{table}

% \section{Baseline Methods}

We compare against four baseline routing strategies: Historical Routes representing actual vessel trajectories extracted from AIS data; Greedy Routing, which myopically selects actions that maximally reduce graph distance to the destination at each step; Dijkstra's, which attempts to reduce the edge cost defined as the edge traversal time + a reliability penalty based on frequency rewards; and A*, which minimizes the same edge cost as Dijkstra's but with an additional distance-based heuristic.

All graph-based baselines use vessel speeds that were calculated as discussed in Section~\ref{subsec:markovian_graph}, where each speed is equal to the average transition speed between adjacent H3-indexed nodes. The performance of the RL agent is obtained by evaluating the final learned models across the six trajectories and three seeds.

To assess generalization, we included three origin--destination pairs that were also used in training and three additional routes chosen to be close to the training ones. This setup tests both reliability on known routes and the ability to adapt to new but related start–goal pairs.

% Temporary
Fig.~\ref{fig:baselines} presents comparative returns across six vessel trajectories in the Gulf of St. Lawrence. Historical routes achieve the lowest average performance, reflecting optimization for objectives not captured in our reward function, such as operational constraints, regulatory requirements, or cargo-specific considerations, rather than poor decision-making. Greedy routing shows moderate performance with notably lower variance, indicating that myopic distance-minimization becomes reasonably stable when operating on a Markovian graph derived from aggregated historical trajectories. A* and Dijkstra achieve higher average returns but with greater variance, confirming that both algorithms effectively exploit the graph structure while remaining sensitive to trajectory-specific characteristics. The RL agent achieves the highest average performance with lower variance across diverse origin--destination pairs.

\color{black}

\section{Conclusion}
\label{sec:conclusion}

% 1. We are only exploring over multi-discrete space, we must explore using continuous action space as thats how autonomous ships are trained.
% 2. We should include physics loss and physics based ...
% 3. Inclusion of more environmental dynamics such as ocean currents, bathymetry, pressure, and what all are important for the ship - for a digital twin.
% 4. Further experimental evaluations such as runnning with SAC, HER (hindsight experience replay)Use Hindsight Experience Replay (HER) to improve learning from sparse rewards in complex goal-driven tasks,.... 
% 5. We are only choosing 3 pairs of similar maritime routes, we need to test it for globally different paths. And also across differnt time periods. We only train on August 2024 wind data.
% 6. Extend MariNav to jointly optimize vessel and port objectives such as multi-port profits and synchronized scheduling.
% 7. Better reward engineering <what all can we say>
% 8. Add more

This paper presented a reinforcement learning framework for maritime navigation that integrates AIS-derived traffic patterns with environmental dynamics on a hexagonal grid. Through the development of the \textit{MariNav} environment, we showed how large-scale vessel tracking data can be transformed into a structured decision-making setting supporting reproducible experimentation. Our evaluation revealed that embedding feasibility constraints through action masking is crucial for training success. At the same time, short observation histories enhance stability across seeds, and intrinsic exploration via RND does not provide benefits under the current weighting scheme. These findings underscore the importance of domain-informed priors, such as traffic graphs and reward shaping, for achieving reliable and interpretable policies.

At the same time, several limitations remain. The current use of a multi-discrete action space restricts alignment with continuous autopilot controls, and the simplified fuel and drag models omit key hydrodynamic processes. Expanding the physical realism of the environment through vessel-specific fuel curves, Nomoto-based dynamics, and differentiable environmental surrogates that include currents, tides, waves, and ice would allow richer and more transferable policies. Algorithmically, reliance on Maskable PPO with RND points to the need for broader evaluation of methods suited to sparse and safety-critical tasks, including Hindsight Experience Replay, constrained RL for COLREGs and ETA compliance, and risk-sensitive formulations such as Conditional Value at Risk.  

Future research should also move beyond single-region evaluation. Broader validation across geographic regions, traffic regimes, and seasonal conditions will be essential to assess global generalization. Multi-agent formulations that model vessel interactions, congestion, and coordination with ports will further enhance realism. Reward engineering should evolve toward potential-based shaping, Pareto optimization, or inverse reinforcement learning to balance efficiency, safety, and compliance in a principled manner. Finally, evaluation metrics must extend beyond episodic return to operational indicators such as ETA deviation, fuel consumption, route similarity, and robustness under counterfactual weather scenarios.  

Taken together, this work establishes a foundation for data-driven reinforcement learning in maritime navigation, providing a configurable and open-source environment for reproducible research. By embedding feasibility constraints, integrating environmental dynamics, and highlighting the role of reward shaping, the proposed framework takes a step closer to becoming a practical decision-support tool for semi-autonomous and autonomous maritime operations.

\section*{Acknowledgment}

This research was partially supported by the \textit{National Council for Scientific and Technological Development} (CNPq 444325/2024-7), the Center for Artificial Intelligence (FAPESP 19/07665-4), the Natural Sciences and Engineering Research Council (NSERC RGPIN-2025-05179), and the Faculty of Computer Science at \textit{Dalhousie University} (DAL). The data used in this study was provided by AISViz/MERIDIAN and is subject to licensing restrictions, preventing the sharing of raw data. However, the pre-trained models and further code can be shared and are available upon request.

\balance
\bibliographystyle{IEEEtran}
\bibliography{IEEEabrv, IEEEreferences}

\begin{thebibliography}{10}
\providecommand{\url}[1]{#1}
\csname url@samestyle\endcsname
\providecommand{\newblock}{\relax}
\providecommand{\bibinfo}[2]{#2}
\providecommand{\BIBentrySTDinterwordspacing}{\spaceskip=0pt\relax}
\providecommand{\BIBentryALTinterwordstretchfactor}{4}
\providecommand{\BIBentryALTinterwordspacing}{\spaceskip=\fontdimen2\font plus
\BIBentryALTinterwordstretchfactor\fontdimen3\font minus \fontdimen4\font\relax}
\providecommand{\BIBforeignlanguage}[2]{{%
\expandafter\ifx\csname l@#1\endcsname\relax
\typeout{** WARNING: IEEEtran.bst: No hyphenation pattern has been}%
\typeout{** loaded for the language `#1'. Using the pattern for}%
\typeout{** the default language instead.}%
\else
\language=\csname l@#1\endcsname
\fi
#2}}
\providecommand{\BIBdecl}{\relax}
\BIBdecl

\bibitem{sharif2023maritime}
N.~Sharif, M.~R{\"o}nnqvist, J.-F. Cordeau, and J.-F. Audy, ``Maritime vessel routing problems with safety concerns: A review,'' 2023.

\bibitem{deraj2023deep}
R.~Deraj, R.~S. Kumar, M.~S. Alam, and A.~Somayajula, ``Deep reinforcement learning based controller for ship navigation,'' \emph{Ocean Engineering}, vol. 273, p. 113937, 2023.

\bibitem{alamoush2025maritime}
A.~S. Alamoush and A.~I. {\"O}l{\c{c}}er, ``Maritime autonomous surface ships: architecture for autonomous navigation systems,'' \emph{Journal of Marine Science and Engineering}, vol.~13, no.~1, p. 122, 2025.

\bibitem{zhang2019decision}
X.~Zhang, C.~Wang, Y.~Liu, and X.~Chen, ``Decision-making for the autonomous navigation of maritime autonomous surface ships based on scene division and deep reinforcement learning,'' \emph{Sensors}, vol.~19, no.~18, p. 4055, 2019.

\bibitem{zhang2025adaptive}
R.~Zhang, X.~Qin, M.~Pan, S.~Li, and H.~Shen, ``Adaptive temporal reinforcement learning for mapping complex maritime environmental state spaces in autonomous ship navigation,'' \emph{Journal of Marine Science and Engineering}, vol.~13, no.~3, p. 514, 2025.

\bibitem{DBLP:journals/corr/abs-1802-09464}
\BIBentryALTinterwordspacing
M.~Plappert, M.~Andrychowicz, A.~Ray, B.~McGrew, B.~Baker, G.~Powell, J.~Schneider, J.~Tobin, M.~Chociej, P.~Welinder, V.~Kumar, and W.~Zaremba, ``Multi-goal reinforcement learning: Challenging robotics environments and request for research,'' \emph{CoRR}, vol. abs/1802.09464, 2018. [Online]. Available: \url{http://arxiv.org/abs/1802.09464}
\BIBentrySTDinterwordspacing

\bibitem{liu2022goal}
M.~Liu, M.~Zhu, and W.~Zhang, ``Goal-conditioned reinforcement learning: Problems and solutions,'' \emph{arXiv preprint arXiv:2201.08299}, 2022.

\bibitem{stach2023maritime}
T.~Stach, Y.~Kinkel, M.~Constapel, and H.-C. Burmeister, ``Maritime anomaly detection for vessel traffic services: A survey,'' \emph{Journal of Marine Science and Engineering}, vol.~11, no.~6, p. 1174, 2023.

\bibitem{michaelides2020decision}
M.~Michaelides, M.~Lind, L.~Green, J.~Askvik, and Z.~Siokouros, ``Decision support in short sea shipping,'' in \emph{Maritime Informatics}.\hskip 1em plus 0.5em minus 0.4em\relax Springer, 2020, pp. 221--236.

\bibitem{dalaklis2023opportunities}
D.~Dalaklis, N.~Nikitakos, D.~Papachristos, and A.~Dalaklis, ``Opportunities and challenges in relation to big data analytics for the shipping and port industries,'' \emph{Smart Ports and Robotic Systems: Navigating the Waves of Techno-Regulation and Governance}, pp. 267--290, 2023.

\bibitem{gao2022mass}
M.~Gao, Z.~Kang, A.~Zhang, J.~Liu, and F.~Zhao, ``Mass autonomous navigation system based on ais big data with dueling deep q networks prioritized replay reinforcement learning,'' \emph{Ocean engineering}, vol. 249, p. 110834, 2022.

\bibitem{bi2024artificial}
J.~Bi, H.~Cheng, W.~Zhang, K.~Bao, and P.~Wang, ``Artificial intelligence in ship trajectory prediction,'' \emph{Journal of Marine Science and Engineering}, vol.~12, no.~5, p. 769, 2024.

\bibitem{drapier2024enhancing}
N.~Drapier, A.~Chetouani, and A.~Chateigner, ``Enhancing maritime trajectory forecasting via h3 index and causal language modelling (clm),'' \emph{arXiv preprint arXiv:2405.09596}, 2024.

\bibitem{zhang2023research}
J.~Zhang, H.~Wang, F.~Cui, Y.~Liu, Z.~Liu, and J.~Dong, ``Research into ship trajectory prediction based on an improved lstm network,'' \emph{Journal of Marine Science and Engineering}, vol.~11, no.~7, p. 1268, 2023.

\bibitem{nguyen2024transformer}
D.~Nguyen and R.~Fablet, ``A transformer network with sparse augmented data representation and cross entropy loss for ais-based vessel trajectory prediction,'' \emph{IEEE Access}, vol.~12, pp. 21\,596--21\,609, 2024.

\bibitem{spadon2024multi}
G.~Spadon, J.~Kumar, D.~Eden, J.~van Berkel, T.~Foster, A.~Soares, R.~Fablet, S.~Matwin, and R.~Pelot, ``Multi-path long-term vessel trajectories forecasting with probabilistic feature fusion for problem shifting,'' \emph{Ocean Engineering}, vol. 312, p. 119138, 2024.

\bibitem{latinopoulos2025marine}
C.~Latinopoulos, E.~Zavvos, D.~Kaklis, V.~Leemen, and A.~Halatsis, ``Marine voyage optimization and weather routing with deep reinforcement learning,'' \emph{Journal of Marine Science and Engineering}, vol.~13, no.~5, p. 902, 2025.

\bibitem{kim2024advancing}
J.~Kim, B.~Hwang, G.-H. Kim, and U.-G. Kim, ``Advancing maritime route optimization: using reinforcement learning for ensuring safety and fuel efficiency,'' \emph{International Journal of e-Navigation and Maritime Economy}, vol.~23, pp. 413--51, 2024.

\bibitem{alam2023ai}
M.~S. Alam, S.~K.~R. Sudha, and A.~Somayajula, ``Ai on the water: applying drl to autonomous vessel navigation,'' \emph{arXiv preprint arXiv:2310.14938}, 2023.

\bibitem{krasowski2024provable}
H.~Krasowski and M.~Althoff, ``Provable traffic rule compliance in safe reinforcement learning on the open sea,'' \emph{IEEE Transactions on Intelligent Vehicles}, 2024.

\bibitem{feng2025safe}
M.~Feng, V.~Parimi, and B.~Williams, ``Safe multi-agent navigation guided by goal-conditioned safe reinforcement learning,'' \emph{arXiv preprint arXiv:2502.17813}, 2025.

\bibitem{spadon2025learning}
G.~Spadon, R.~Song, V.~Vaidheeswaran, M.~M. Alam, F.~Goerlandt, and R.~Pelot, ``Learning spatio-temporal vessel behavior using ais trajectory data and markovian models in the gulf of st. lawrence,'' \emph{arXiv preprint arXiv:2506.00025}, 2025.

\bibitem{sutton2018reinforcement}
R.~S. Sutton and A.~G. Barto, \emph{Reinforcement Learning: An Introduction}, 2nd~ed.\hskip 1em plus 0.5em minus 0.4em\relax The MIT Press, 2018.

\bibitem{schulman2017proximal}
J.~Schulman, F.~Wolski, P.~Dhariwal, A.~Radford, and O.~Klimov, ``Proximal policy optimization algorithms,'' \emph{arXiv preprint arXiv:1707.06347}, 2017.

\bibitem{burda2018exploration}
Y.~Burda, H.~Edwards, A.~Storkey, and O.~Klimov, ``Exploration by random network distillation,'' \emph{arXiv preprint arXiv:1810.12894}, 2018.

\bibitem{rahayu2024optimizing}
D.~A. Rahayu, I.~Ardiyanto, W.~Hasbi \emph{et~al.}, ``Optimizing maritime vessel trajectory prediction using space-based ais data and pso-bigru,'' in \emph{2024 IEEE International Conference on Aerospace Electronics and Remote Sensing Technology (ICARES)}.\hskip 1em plus 0.5em minus 0.4em\relax IEEE, 2024, pp. 1--7.

\bibitem{huang2020closer}
S.~Huang and S.~Onta{\~n}{\'o}n, ``A closer look at invalid action masking in policy gradient algorithms,'' \emph{arXiv preprint arXiv:2006.14171}, 2020.

\bibitem{Brodsky_H3_Uber}
\BIBentryALTinterwordspacing
I.~Brodsky. (2018, 08) H3: Uber’s hexagonal hierarchical spatial index. Uber Engineering Blog. Posted by Isaac Brodsky. Uber Engineering. [Online]. Available: \url{https://www.uber.com/en-CA/blog/h3/}
\BIBentrySTDinterwordspacing

\bibitem{spadon2024maritime}
G.~Spadon, J.~Kumar, J.~Chen, M.~Smith, C.~Hilliard, S.~Vela, R.~Gehrmann, C.~DiBacco, S.~Matwin, and R.~Pelot, ``Maritime tracking data analysis and integration with aisdb,'' \emph{SoftwareX}, vol.~28, p. 101952, 2024.

\bibitem{hersbach2020era5}
H.~Hersbach, B.~Bell, P.~Berrisford, S.~Hirahara, A.~Hor{\'a}nyi, J.~Mu{\~n}oz-Sabater, J.~Nicolas, C.~Peubey, R.~Radu, D.~Schepers \emph{et~al.}, ``The era5 global reanalysis,'' \emph{Quarterly journal of the royal meteorological society}, vol. 146, no. 730, pp. 1999--2049, 2020.

\bibitem{Kim2020ISO15016Fuel}
T.~Kim, J.-I. Hwang, S.-H. Kim, D.-G. Park, and J.-H. Seo, ``Iso 15016:2015-based method for estimating the fuel oil consumption of a ship,'' \emph{Journal of Marine Science and Engineering}, vol.~8, no.~10, p. 791, 2020.

\bibitem{Zhang2020AirResistanceContainerShip}
Z.~Zhang, S.~Sun, S.~Wang, and Z.~Liu, ``Comparative study of air resistance with and without a superstructure on a container ship using numerical simulation,'' \emph{Journal of Marine Science and Engineering}, vol.~8, no.~4, p. 267, 2020.

\bibitem{Wang2019ObliqueWindModel}
Q.~Wang, C.~Guo, S.~Yang, and Y.~Wang, ``Numerical simulation of container ship in oblique winds to develop a wind resistance model based on statistical data,'' \emph{Ships and Offshore Structures}, vol.~14, no. sup1, pp. S285--S296, 2019.

\bibitem{Zhang2021SpeedOptimizationFleet}
R.~Zhang, J.~Wang, and Q.~Meng, ``Speed optimization for container ship fleet deployment considering fuel consumption,'' \emph{Sustainability}, vol.~13, no.~9, p. 5242, 2021.

\bibitem{stable-baselines3}
\BIBentryALTinterwordspacing
A.~Raffin, A.~Hill, A.~Gleave, A.~Kanervisto, M.~Ernestus, and N.~Dormann, ``Stable-baselines3: Reliable reinforcement learning implementations,'' \emph{Journal of Machine Learning Research}, vol.~22, no. 268, pp. 1--8, 2021. [Online]. Available: \url{http://jmlr.org/papers/v22/20-1364.html}
\BIBentrySTDinterwordspacing

\end{thebibliography}

\end{document}